%% file: main.tex
\pdfoutput=1

\documentclass[11pt]{article}

\usepackage{acl}

\usepackage{times}
\usepackage{latexsym}
\usepackage{amsmath}
\usepackage{amssymb}
\usepackage{algorithm}
\usepackage{algpseudocode}

\algtext*{EndWhile}
\algtext*{EndIf}
\algtext*{EndFor}
\usepackage{graphicx}
 \usepackage{array,multirow}
 \usepackage{booktabs}

\usepackage[T1]{fontenc}

\usepackage[utf8]{inputenc}

\usepackage{microtype}

\usepackage{bbm}
\usepackage{xurl}

\include{aux}
\newcommand{\naver}{$^1$}
\newcommand{\ind}{$^2$}

\title{Should you marginalize over possible tokenizations?}

\author{ Nadezhda Chirkova\naver~\; Germ\'an Kruszewski\naver~\; Jos Rozen\naver~\; Marc Dymetman\ind~\;\\
  \naver Naver Labs Europe \quad \ind Independent Researcher \\
  \texttt{\{nadia.chirkova, german.kruszewski, jos.rozen\}@naverlabs.com} \\
  \texttt{marc.dymetman@gmail.com}\\
  }

\begin{document}
\maketitle
\begin{abstract}
Autoregressive language models (LMs) map token sequences to probabilities.
The usual practice for computing the probability of any character string (e.g. English sentences) is to first transform it into a sequence of tokens that is scored by the model.
However, there are exponentially many token sequences that represent any given string.
To truly compute the probability of a string one should \emph{marginalize} over all tokenizations, which is typically intractable.
Here, we analyze whether the practice of ignoring the marginalization 
is justified.
To this end, we devise an importance-sampling-based algorithm that allows us to compute estimates of the marginal probabilities and compare them to the default procedure in a range of state-of-the-art models and datasets.
Our results show that the gap in log-likelihood
is no larger than 0.5\% in most cases, but that it becomes more pronounced for data with long complex words.
\end{abstract}

\section{Introduction}

Language models are probability distributions over text strings.
In practice, these distributions are defined over a vocabulary of \emph{tokens}, such as
words, punctuation marks, and other special symbols~\citep{Jurafsky:2000, Goldberg:2017}.
As long as a unique token sequence encodes any given string, the probability of a string according to the language model is equal to the probability of the corresponding token sequence.
However, with today popular sub-word-level tokenizations this is not the case, as there are
(exponentially) many possible tokenizations for any given string. 
For example, with the vocabulary $V=\{a, ab, b, c, ca, cab\}$, the string \textit{``cab''} can be tokenized into ${cab, c/a/b, ca/b, c/ab}$.
Therefore, the \emph{true} probability that the language model assigns to the corresponding string is that obtained after marginalizing over \emph{all possible tokenizations}. 
Yet, the common practice disregards this fact, computing the string probability by scoring a single \textit{default} tokenization (e.g., $cab$). 
The implicit assumption from the community is that the probability mass of non-default tokenizations is negligible.
However,
this assumption has not been adequately evaluated yet.

\begin{figure}[t!]
    \centering
        \centering
        \includegraphics[width=\linewidth]{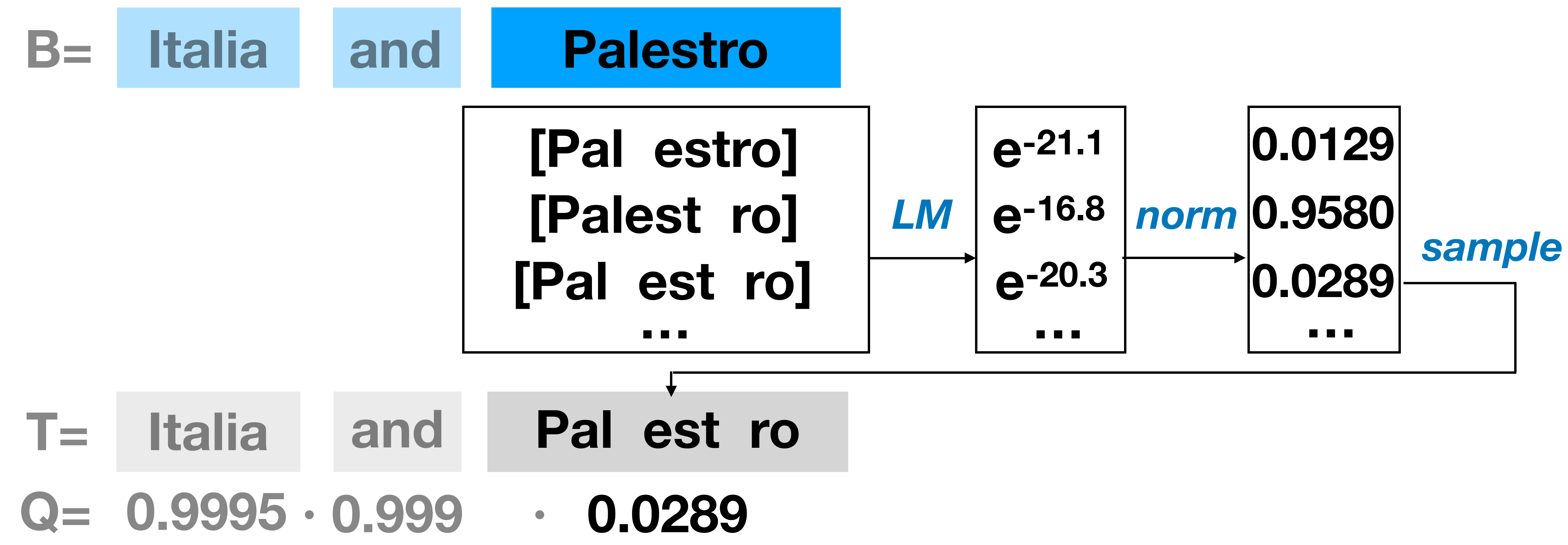}
        \caption{Illustration of the proposed procedure for sampling tokenization $T$ and calculating its proposal probability $Q=Q(T|S)$ from a sequence of blocks $B$, produced by splitting sequence $S$.}
        \label{fig:ill}
\end{figure}

In part,
\citet{cao-rimell-2021-evaluate} addressed this very same question, 
by conducting
a pioneer study to quantify the gap between the default and marginalized probabilities.
Their experiments with Transformer-XL pretrained on the WMT data (English and German) show negligible changes in perplexity with respect to using a single default tokenization for in-domain data and 0.9--1.9\% improvement in perplexity for out-of-domain data, such as arXiv articles. 
Because exact marginalization is intractable in practice, marginalized probabilities were estimated using importance sampling.
Importance sampling computes an unbiased estimate of the marginalized probabilities 
as an average over
tokenizations sampled from a proposal distribution.
\citet{cao-rimell-2021-evaluate} exploited the probabilistic nature of the UnigramLM tokenizer~\citep{Kudo:2018} to define such a proposal.
As a consequence, their results do not necessarily
extend to the more popular language models like GPT\ptjr{-}2~\citep{Radford:etal:2019}, GPT\ptjr{-}3~\citep{Brown:etal:2020}, BLOOM~\citep{BLOOM}, T5~\citep{Raffel:etal:2020}, 
among others, trained using other tokenization schemes such as BPE~\citep{Sennrich:etal:2016}, WordPiece~\citep{Schuster:Nakajima:2012}, among others.

In this work, 
we devise a new proposal distribution that allows us to quantify the effect of marginalization for any given tokenizer.
Equipped with this algorithm, we inspect the effect of marginalization over tokenizations for two LMs, GPT-2 (126M parameters, English) and the recently released BLOOM (1.7B parameters, multilingual), on various domains and languages.
Our importance sampling estimates show that in practice marginalization does not influence log-likelihood much (usually less than 0.5\% improvement), 
the highest influence (1--2\% improvement) being for data with long, complex words and distribution shift.
Because the results will vary for different models and data, we 
provide a tool for researchers and practitioners to measure the gap 
in their specific setting to decide whether the usual practice is warranted.
To this end, we release our code\footnote{ \url{https://github.com/naver/marginalization}}, which
can be applied to models from the \texttt{transformers} library.

\section{Methodology}
\subsection{Preliminaries}
Let us consider a sequence of characters $S$ that we wish to score with an autoregressive language model $\Ptheta$.
Typically, $S$ is split into a sequence $T=t_1, \dots, t_n$ of tokens $t_i\in V$, where $V$ is the model's vocabulary, 
a process
commonly known as \emph{tokenizing} the sequence.
Then we can compute a score for a tokenization $T$ of the sequence $S$, 
$\Ptheta(T, S)$,
using the chain rule: $$\Ptheta(T, S) = \mathbbm{1}[T\rightarrow S]\prod_{j=1}^{|T|} \Ptheta(t_j|t_{j-1}, \dots, t_{1})$$ where $T\rightarrow S$ indicates that $T$ is a valid tokenization of $S$. 
Commonly used tokenization algorithms such as BPE or WordPiece provide a deterministic procedure for obtaining a 
\emph{particular}
way of tokenizing $S$ into $T$, which we refer to as the \emph{default} tokenization. 
Yet, in general, for the same sequence, there exist (exponentially) many possible tokenizations with vocabulary $V$, which also typically receive some probability mass by the LM.
To obtain the \emph{true} probability score for the sequence $S$, we should marginalize over all valid tokenizations: 
$\Ptheta(S) = \sum_{T: T\rightarrow S} \Ptheta(T, S)$. 

However, computing $\Ptheta(S)$ is typically intractable given the exponential number of valid tokenizations. Nonetheless, this value can be estimated through importance sampling, as follows.
Introducing a proposal distribution $Q(T|S)$ over all tokenizations $T$ of a sequence $S$, such that $P(T,S) > 0 \Rightarrow Q(T|S) > 0$, we can rewrite the probability
$P(S)$, as follows:
\begin{equation}
    \!\Ptheta(S) = \sum_{T: T\rightarrow S} \Ptheta(T, S) = \mathbb{E}_{Q(T|S)} \frac {\Ptheta(T, S)}{Q(T|S)}
\end{equation}
Now we can estimate $\Ptheta(S)$  by sampling $K$ independent tokenizations from the proposal:
\begin{equation}
    \label{eq:is}
    \Ptheta(S) \approx \frac 1 K \sum_{k=1}^K \frac {\Ptheta(T_k, S)}{Q(T_k|S)}, \quad T_k \sim Q(T|S)
\end{equation}
The quality of this estimate depends on the chosen proposal distribution: the closer the proposal $Q(T|S)$ is to the true posterior distribution $P(T|S)$, the smaller the variance of the unbiased estimate~\eqref{eq:is} tends to be.\footnote{If we had access to the true posterior distribution $P(T|S)$,
we would have $\frac{P(T, S)}{P(T|S)} = P(S)$, and therefore
(i) one sample would be enough to obtain the needed value $P(S)$, and (ii) the variance of the importance sampling estimate would be zero.\label{ft:true-posterior}}

\begin{algorithm}[t]
\caption{Proposal algorithm}\label{alg:sampling}
\begin{small}
\begin{algorithmic}[1]
\Require sequence $S$; max. block size $L$; max. number of tokenizations per block $M$
\Ensure a tokenization $T$ sampled with prob. $Q(T|S)$
\State $T \gets [\,]$; $q \gets 1$
\State $B \gets \mathrm{split\_in\_blocks}(S, L)$
\For{$i = 1, \dots, |B|$}
    \State $X \gets \mathrm{get\_all\_tokenizations}(B_i, M)$
    \For{$j = 1, \dots, |X|$}
        \State $\hat s_j \gets \mathrm{LM}(X_j|T)$
    \EndFor
    \For{$j = 1, \dots, |X|$}
        \State $s_j = \hat s_j / \sum_{j} \hat s_j$
    \EndFor
    \State $j_* \gets \mathrm{sample}(s_1, \dots, s_{|X|})$
    \State $T \gets \mathrm{concat}(T, X_{j_*})$
    \State $q \gets q \cdot s_{j_*}$
\EndFor
\State $Q(T|S) \leftarrow q$
\State \textbf{return} $T$, $Q(T|S)$
\end{algorithmic}
\end{small}
\label{al:proposal}
\end{algorithm}

\subsection{Proposed approach}
\label{sec:approach}

We introduce a novel
proposal $Q(T|S)$ based on the LM itself 
 with the intention to make it naturally closer 
to the posterior.
Importantly, this proposal can be used for \emph{any} tokenizer
enabling its application to well-known state-of-the-art systems.
The procedure for sampling
from this proposal is presented in Algorithm \ref{al:proposal} and also illustrated in Figure \ref{fig:ill}.
In summary, the algorithm samples a tokenization $T$ by building it incrementally as the concatenation of token subsequences $T_i$. 
Each token subsequence is sampled from the language model 
while always ensuring that the resulting tokenization is valid for the target $S$. 
To achieve this, the algorithm breaks $S$ into a sequence of character blocks $B$, and \emph{only} samples tokenizations 
$T_i$ \emph{that are valid for the corresponding block  $B_i$}.
Notably, in the extreme case of splitting $S$ into 
a single block $B_1=S$, 
our proposal $Q(T|S)$ turns into the true posterior $P(T|S)$, allowing to compute the exact marginalization with a single sample, as noted in footnote \ref{ft:true-posterior}.
However, because sampling a valid tokenization of a block requires renormalizing over \emph{all} such valid tokenizations, 
this 
extreme
instantiation 
would defeat the purpose of the algorithm as it would be equivalent to 
computing the full marginalization.
Instead, we consider block sizes over which we can practically compute the renormalization constant 
by, for example, using whitespace-separated words as blocks.
Still, because this can sometimes lead to impractically-sized blocks 
with a 
number of tokenizations 
that can exceed what we can reasonably score with a LM, 
we limit the maximum block size to a parameter $L$ and we only score 
the top $M$ block tokenizations inversely sorted by their number of tokens%
\footnote{Appendices \ref{app:maxblocklength} and~\ref{app:maxtokblock} describe practical details of these hyperparameters.}.
The resulting algorithm requires $O(|B| \times M)$ evaluations of the LM per-sample, where $|B|$ is the number of blocks used to split the sequence $S$.
In Appendix~\ref{app:valid}, we validate 
that, for short sentences with a tractable amount of possible tokenizations, 
for which we can actually compute the true value of the marginalization,
our algorithm 
provides quite precise estimates.

\section{Experiments}
\label{sec:exp}

\begin{table}[t]
\centering
\begin{small}
\begin{tabular}{p{0.95cm}|ccp{0.8cm}p{0.8cm}p{0.4cm}}
 \toprule
Data &  $\BPCst$ &  $\BPCis$ & BPC gap & \% rel. gap & \% ND \\
\midrule 
\multicolumn{6}{c}{GPT-2 (125M params)} \\ \hline
                 Wiki & 1.1076 & 1.1026 &        .0050 &    0.45\% &             0.9\% \\
              Twit & 1.9610 & 1.9303 &        .0307 & 1.56\%    &             4.2\% \\ 
              News & 0.9421 & 0.939 &        .0028 & 0.30\%    &             0.4\% \\
              Tr.sp. & 1.0234 & 1.0029 &      .0204 & 1.99\%    &             1.5\% \\ \hline
\multicolumn{6}{c}{BLOOM (1.7B params)} \\ \hline
              Twit & 1.7889 & 1.7653 &        .0236 &  1.32\%   &             3.3\% \\
        News & 0.8499 & 0.8462 & .0037 &  0.55\%       & 0.4\%                 \\
        Tr.sp. & 0.9022 & 0.9002 &  .0020       & 0.23\%   &  0.4\%            \\
          Chi$^\dagger$ & 1.2080 & 1.2024 &        .0056 &    0.46\%      &        3.1\%  \\
          Fra & 0.8001 & 0.7993 &        .0008 &   0.10\%       &        0.2\%   \\ 
              Spa & 0.8813 & 0.8800 &        .0013 &  0.14\%    &            0.3\%  \\
           Vie & 0.7939 & 0.7932 &        .0008 &    0.10\%    &          0.1\%   \\
           Ind & 0.9812 & 0.9778 &        .0034 &    0.34\%     &         0.6\%  \\
            Eus & 1.2432 & 1.2269 &        .0163 &  1.31\%   &             3.5\%  \\
            Urd$^\dagger$ & 0.8785 & 0.8697 &        .0088 &    1.00\%    &          1.8\%  \\
        Python & 0.5100 & 0.5071 &        .0029 &  0.56\%     &           1.3\%   \\
        C++ & 0.6053 & 0.5993 &       .0059 &  0.98\%     &           2.2\%   \\
\bottomrule
\end{tabular}
\end{small}
\caption{\label{tab:main}
Main results. All natural languages from Flores-200 devtest set, sorted decreasingly by the size of corpora\textsuperscript{*} used in BLOOM's training. $^\dagger$ denotes non-latin script languages. }
\begin{flushleft}
\footnotesize\textsuperscript{*} Corpora sizes available at \url{https://huggingface.co/bigscience/bloom#training-data}
\end{flushleft}
\end{table}

\paragraph{Experimental setup.}
We experiment with two language models, GPT-2 (\citealt{Radford:etal:2019}, 126M parameters, English) and the recently released BLOOM (\citealt{BLOOM}, 1.7B parameters, 45 natural and 12 programming languages). We 
select
the following datasets for evaluating the LMs, which cover different styles and languages: Wikipedia articles (En), Twitter posts (En), CNN news (En), Transcriptions of White House Speeches (En), Flores-200 (sentences from Wikipedia in various languages, including high-resource, low-resource, latin and non-latin scripts), Python and C++ code (one recently released repository for each language).
We concatenate texts into sequences of length 800 tokens (as measured by the default tokenization) to provide longer context for the LM. We evaluate on 100 sequences per dataset (Flores-200, CNN news and Code datasets are shorter). We refer to Appendix~\ref{app:data} for more details on the data and how we check that the LMs were not trained on the evaluation data.

We measure the cross entropy (in BPC\footnote{Bits Per Character: $\BPC=-(\log_2 \mathrm{Prob}(S)) / |S|$, where $\mathrm{Prob}$ is the probability assigned to $S$: $P(T, S)$ in $\BPCst$ or $P(S)$ in $\BPCis$.}) between the data and the model according to the default tokenization ($\BPCst$) and between the data and the marginalized model according to the importance sampling estimate ($\BPCis$), as well as their difference $\BPCst - \BPCis$ referred to as the BPC \textit{gap}, and also the normalized difference $(\BPCst - \BPCis)/\BPCst$ (relative BPC gap).
Furthermore, we compute a 90\% confidence interval $[\BPCisL, \BPCisR]$ around $\BPCis$,
using bootstrap resampling~\citep[Chapter 8]{Wasserman:2004} for $n=1000$ trials\footnote{We use the \href{https://docs.scipy.org/doc/scipy/reference/generated/scipy.stats.bootstrap.html}{scipy.stats.bootstrap} implementation, with \texttt{method='BCa'}.}.
Additionally, we report the proportion of blocks for which our algorithm 
samples
non-default tokenizations (\%ND).

As for hyperparameters, we use $M=128$ 
and choose $L$ to be
the maximum token length in the default tokenization of the evaluation data. 
We provide empirical validation for both 
these hyperparameters in Appendices \ref{app:maxtokblock} and  \ref{app:maxblocklength}, respectively.
We sample $K=30$ tokenizations per sequence.

\paragraph{Results} Table~\ref{tab:main} presents our main results.
We generally observe a low relative BPC gap ($<0.5\%)$, 
but in some cases exceeding 1\%,
e.g. 1.3--1.5\% on Twitter, 2\% on transcribed speech data, 1.3\% on the Basque language (Eus) or 1\% on the Urdu language (Urd). 
We note that dataset/model pairs with higher relative gap tend to be 
connected with low-resource languages (Basque and Urdu), non-latin scripts (Urdu and Chinese), and data distribution shift (transcribed speech, Twitter). Moreover, we observe a higher gap to be associated with a higher percentage of non-default tokenizations sampled by our algorithm (\%ND). 
To learn more about the factors driving the probability of sampling the default tokenization, we bin blocks (which roughly correspond to words) from Wikipedia by the probability that our proposal assigns to 
their 
default tokenization, $Q(\mathrm{df.})$, when using GPT-2 as a model. Table~\ref{tab:examples} shows a few examples of blocks from each bin alongside the bin's frequency.
As can be seen, high probability of sampling the default tokenization usually corresponds to common and simple words, whereas low probability
corresponds to complex and rare words. 
From this observation, we conjecture 
that higher gaps are at least in part driven by the presence of long complex words in the datasets.

\begin{figure}[t!]
    \centering
        \centering
        \includegraphics[width=\linewidth]{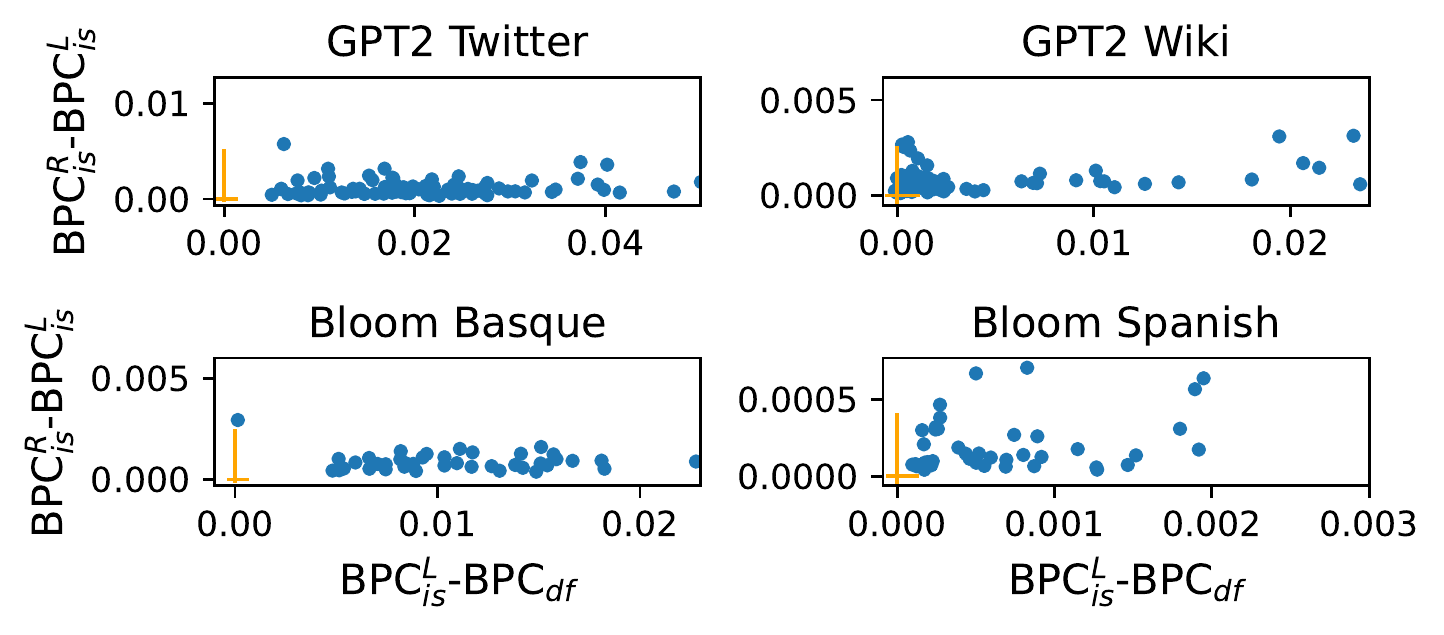}
        \caption{Confidence intervals visualisation. Each dot represents one data point (sequence).}
        \label{fig:confints}
\end{figure}

\begin{table}[t!]
\begin{small}   
\centering
\begin{tabular}{p{0.9cm}|c|p{4.6cm}} 
\toprule
$Q$(df.) & Freq. & Example blocks\\ \midrule
>0.999 & 90\%& Many, are, the, larger, amphibians, superficially, resemble \\ \hline
0.99--0.999 & 6.1\%& crocodiles, whenever, bases, Rifenburg, sailed, precursors \\  \hline
0.9--0.99 & 2.2\%& warships, propelled, Tomasz, redemption, Metoposaurus \\  \hline
0.5--0.9 & 0.7\% & paedomorphic, Peltobatrachus, ironclad, Urabi, Tonnante  \\  \hline
0--0.5 & 0.7\% & temnospondyls, brevirostrine, Pugong, saurus,
semiaquatic\\ 
\bottomrule
\end{tabular}
\end{small}
\caption{Examples of blocks 
binned by
proposal probability of the default tokenization, with percentage of such blocks in the dataset. GPT-2 on Wikipedia data.}
\label{tab:examples}
\end{table}

Finally, 
Figure~\ref{fig:confints} visualizes confidence intervals on BPC gaps 
 for \emph{individual} sequences across several datasets. Additional results are given in Appendix~\ref{app:confints}. 
 In particular, we plot the left limit of the confidence interval for
 the BPC gap ($\BPCisL-\BPCst$) on the $x$-axis and the width of the interval ($\BPCisR-\BPCisL$) on the $y$-axis (non-negative by definition). If a dot is located to the right of 0, it means that 
 we are highly confident that the BPC gap is positive on that individual sequence.
 The farther the dot is on the $x$-axis, the higher the corresponding BPC gap is.
 Likewise, the lower the value 
 on the $y$-axis, the lower is the variance of our 
 estimate of the marginalized probability and, consequently, of the BPC gap.
 As 
 can be seen, we 
 obtain
 low-variance predictions for most of the sequences, and for almost all of them we can observe a positive BPC gap. 
 Moreover, we can 
 note a
 distributional difference between dataset/model pairs with a low BPC gap (such as those on the right-hand side of Figure \ref{fig:confints}, with points concentrated close to the 0 value) and those with high BPC gap (such as those represented on the left-hand side of Figure \ref{fig:confints}, with points spread up to the right).

\section{Related Work}
Stochastic tokenization or marginalisation over tokenizations were widely investigated in the context of model \textit{training}~\citep{grave-hybrid, Merrienboer, buckman-neubig-2018-neural, bpe-dropout, Kudo:2018} or learning better tokenizers~\cite{he-dynamic}; in contrast, we evaluate the effect of marginalization at the \textit{inference} stage, when the tokenizer and the LM were trained in the default, commonly-used way. 
The closest study to ours is \citet{cao-rimell-2021-evaluate}, which 
relies on the stochastic version of the UnigramLM tokenizer as their proposal $Q$, and thus their approach is inapplicable to LMs with other tokenizers. They also had to introduce a set of heuristics such as 
imposing consistent tokenization of repeated words or enforcing the default tokenization to be included among the sampled tokenizations, to make this proposal closer to the posterior and to decrease the variance of importance sampling.

\section{Conclusion}
In this work, we have studied the effect of marginalization over possible tokenizations in language modeling. For this, we introduced
a novel proposal distribution over tokenizations, which is used in the importance sampling algorithm to obtain estimates of the marginalized probability, and that can be applied to any tokenizer and language model. Our results show that the overall effect of marginalization over tokenizations is often smaller than 0.5\%, although it becomes more pronounced for data with long complex words or distribution shift. We release
our code to allow practitioners to check the effect of marginalization for their models of interest.

\section*{Limitations}
The main limitation of the proposed approach is that it would be relatively costly to apply at production time, compared to the conventional LM evaluation.
First, it requires drawing a number of tokenization samples, as defined by importance sampling, in contrast to a single pass through the evaluated sequence in the conventional approach. Second, the conventional approach can be conducted with teacher forcing and efficiently parallelized, while the proposed approach relies on block-by-block sequential processing. Nonetheless, the proposed algorithm is designed for analysis purposes rather than to be used in production systems, for which it is feasible to run it in a reasonable time, allowing users to evaluate the effect of marginalization for any tokenizer and language.

\section*{Broader impact}
As the work is dedicated to evaluating existing models on publicly available datasets, we are not aware of any potential negative impact.

\section*{Acknowledgements} 
We would like to thank Matthias Gall\'e for his valuable feedback.

\bibliography{anthology,custom} %
\bibliographystyle{acl_natbib}

\clearpage
\appendix

\section{Data}
\label{app:data}

We consider the following datasets: 
\begin{itemize}
    \item Wikitext (\url{https://huggingface.co/datasets/wikitext}, \verb|wikitext-2-raw-v1| test subset, \citealp{merity2016pointer}, CC BY-SA 4.0 license);
    \item Twitter posts (\url{https://huggingface.co/datasets/tweet_eval}, \verb|emoji| test subset, \citealp{mohammad2018semeval});
    \item CNN News (\url{https://www.kaggle.com/datasets/hadasu92/cnn-articles-after-basic-cleaning}, CC0 license);
    \item The White House Speeches (\url{https://www.kaggle.com/datasets/mohamedkhaledelsafty/the-white-house-speeches-and-remarks-12102022}, CC0 license);
    \item Flores-200 (\url{https://github.com/facebookresearch/flores/tree/main/flores200}, \citealp{nllb2022}, CC BY-SA 4.0 license);
    \item Python Code (all \verb|.py| files from \url{https://github.com/naver/disco}, Creative Commons Attribution-NonCommercial-ShareAlike 4.0 license);
    \item C++ Code (all \verb|.h| and \verb|.cc| files from \url{https://github.com/microsoft/Trieste}, MIT license).
\end{itemize}
Wikitext and White House Speeches datasets consist of paragraphs extracted from Wikipedia articles (\url{wikipedia.org}) or from transcribed speeches. 
Flores-200 is composed of sentences extracted from English Wikipedia and translated by professional translators into 200 languages.  
Python and C++ Code data consists of code files. Twitter / News datasets consist of separate tweets / news articles.
We compose sequences to evaluate an LM on, by concatenating texts listed above into sequences of 800 tokens according to the
default tokenization (concatenated texts are separated by \verb|\n\n|). The sequence always begins with a new text. Code and News data contains texts longer than 800 tokens, these texts are considered as separate sequences and clipped to 800 tokens.  Table~\ref{tab:data} reports statistics of the data. Maximum 100 sequences per dataset are considered (Flores-200 dataset, News dataset and code data are shorter).

We checked that the data we evaluate on was not used in model training as follows. GPT-2 was not trained on Wikipedia data, as reported in its paper~\cite{Radford:etal:2019}.  BLOOM was trained on Wikipedia data, so we do not evaluate it on Wikipedia and English Flores data. At the same time, data for other languages is based on translations, which makes it safe to use it for evaluation. 
Twitter is not listed in data sources for GPT-2 (\url{https://github.com/openai/gpt-2/blob/master/domains.txt}) and BLOOM (\url{https://huggingface.co/spaces/bigscience/BigScienceCorpus}). 
For evaluation on code, we use the code of the libraries created after the BLOOM's training. Likewise, for evaluation on the news and White House speech data, we selected only texts released after 11.03.2022 (after the beginning of the largest BLOOM model's training).

\section{Additional information on experiments}
The BLOOM model is released under the Responsible AI License, and GPT-2 is released under the Modified MIT License. Our code is based on the \texttt{transformers} library~\cite{wolf-etal-2020-transformers} which is released under the Apache License 2.0 license. All assets allow usage for research purposes. Evaluation of the GPT-2 model was conducted on a single Tesla V-100 GPU (24--48 GPU hours per dataset), and evaluation of the BLOOM model conducted on a single Tesla A100 GPU (72--120 GPU hours per dataset).

\begin{table}[t!]
\centering
\begin{tabular}{p{2cm}|p{2cm}|p{2cm}} 
\toprule
\textbf{Dataset} & \textbf{Av.~/~max length} & \textbf{Total \# of sequences} \\
 \midrule
Wikitext & 98~/~556  & 100  \\
Twitter &  20~/~159  & 100  \\
News &  833~/~2940  & 63  \\
Tr. sp. &  33~/~158  & 100  \\
Flores (En) &  27~/~69  & 37  \\
Python &  320~/~2623  & 6  \\
C++ &  2296~/~16324  & 12  \\
\bottomrule
\end{tabular}
\caption{Data statistics. Average and maximal length: length of BLOOM's tokenization for raw texts from the dataset (before concatenation). For Flores we only list English as an example, as other language data is a translation of English sentences.}
\label{tab:data}
\end{table}

\section{Segmentation into blocks}
\label{app:maxblocklength}
\begin{figure}[t!]
    \centering
        \centering
        \includegraphics[width=0.8\linewidth]{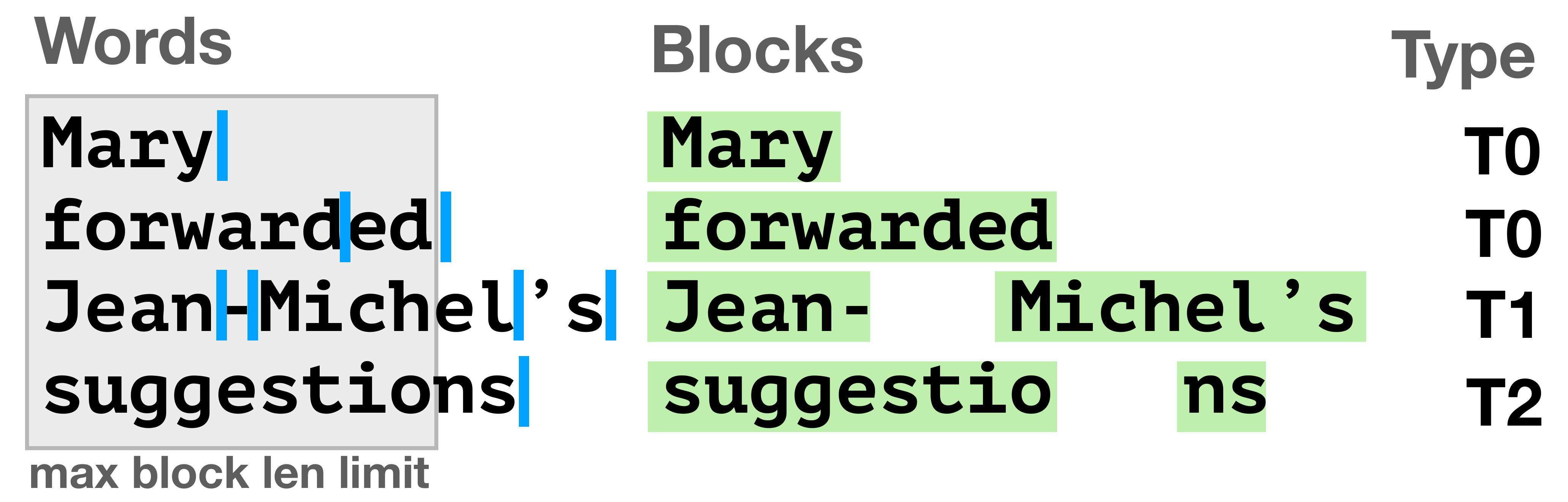}
        \caption{Illustration of blocks composition from whitespace- or newline-separated words. Vertical blue bars denote default tokenization. T1 and T2 blocks occur for words longer than the maximum block length $L$, here equal to 9.}
        \label{fig:blocks}
\end{figure}

As discussed in Section \ref{sec:approach}, the proposal algorithm splits the sequence into a sequence of blocks.
In our experiments, we split the sequence at white spaces and new line characters, thus making blocks roughly correspond to words.
Because our algorithm computes all possible tokenizations within a block, this process can become prohibitively expensive for long blocks, which can occur with complex words or in languages that do not frequently use the white space character, such as Chinese.
For this reason, 
we define a \textit{maximum block length} hyperparameter, $L$.
Words that have length lower or equal to $L$ are denoted as type 0 (T0) blocks.
If a word has length larger than $L$, it is split into smaller blocks, as follows.
First, we compute the block's default tokenization and incrementally merge the tokens while checking not to exceed $L$. 
Once the limit is reached, a new block is started.
The resulting blocks are denoted as type 1 (T1) blocks.
Suppose at any point a token of length larger than $L$ is encountered. In that case, this token is cropped at $L$, and the remaining characters are then moved to a new block.
These blocks are denoted as type 2 (T2) blocks.
Figure~\ref{fig:blocks} illustrates these three block types.

\begin{table}[t!]
\centering
\begin{small}
\begin{tabular}{c|c|ccccc} 
\toprule
& $L$ & \%T2 & \%T1 & BPC gap & \% $\BPCis<\BPCst$ \\ \midrule
\parbox[t]{2mm}{\multirow{3}{*}{\rotatebox[origin=c]{90}{French}}} & 17 &0.08 & 0.5 & -0.00108 & 77 \\
& 19 & 0 & 0.33 & 0.00084 & 100 \\
& 21 & 0 & 0.08 & 0.00079 & 98 \\ \hline
\parbox[t]{2mm}{\multirow{3}{*}{\rotatebox[origin=c]{90}{Urdu}}} & 19 & 0.3 & 1.3 & 0.0019 & 62\\
& 21 & 0 & 0.3 & 0.0087 & 100\\
& 23 & 0 & 0.2 & 0.0089 & 100\\ \bottomrule
\end{tabular}
\end{small}
\caption{Effect of the maximum block length hyperparameter, $L$, on the portion of T1 and T2 blocks, BPC gap and the portion of sequences with $\BPCis<\BPCst$. Model: BLOOM-1.7B.}
\label{tab:mbl}
\end{table}

Table~\ref{tab:mbl} illustrates the effect that the maximum block length hyperparameter $L$ has for BLOOM on French (low-gap case) and Urdu (higher-gap case). 
We experiment with three values of $L$ to represent various proportions of T1 and T2 blocks. 
For low values of $L$ ($L=17$ and $L=19$ for French and Urdu, respectively), we observe some small or even negative gap in BPC, and a large percentage of sequences that have higher cross-entropy when using the marginal than when using the default tokenization.
This result comes with a small but non-negligible percentage of T2 blocks.
Because T2 splits a token that is selected by the default tokenization across different blocks, this prevents the proposal from ever sampling the default tokenization, resulting in a poor estimate.
Higher 
values of $L$ result in the elimination of any T2 blocks with also a moderate impact on T1 blocks.
Yet, once T2 blocks are eliminated, the number of T1 blocks does not appear to have a sizeable effect.
Overall, these results provide the rule for selecting $L$: it should be set to the maximum length of the tokens in the default tokenization of the evaluation data in order to avoid T2 blocks.%

\section{Limiting the number of tokenizations per block}
\label{app:maxtokblock}
\begin{figure}[t!]
    \centering
        \centering
        \includegraphics[width=\linewidth]{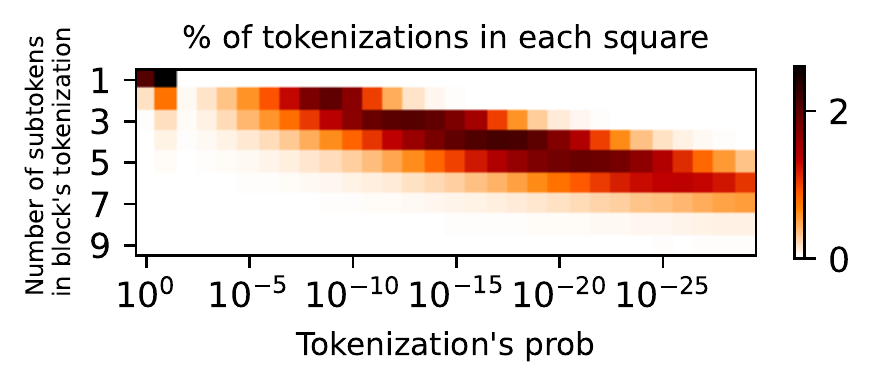}\\
        \includegraphics[width=\linewidth]{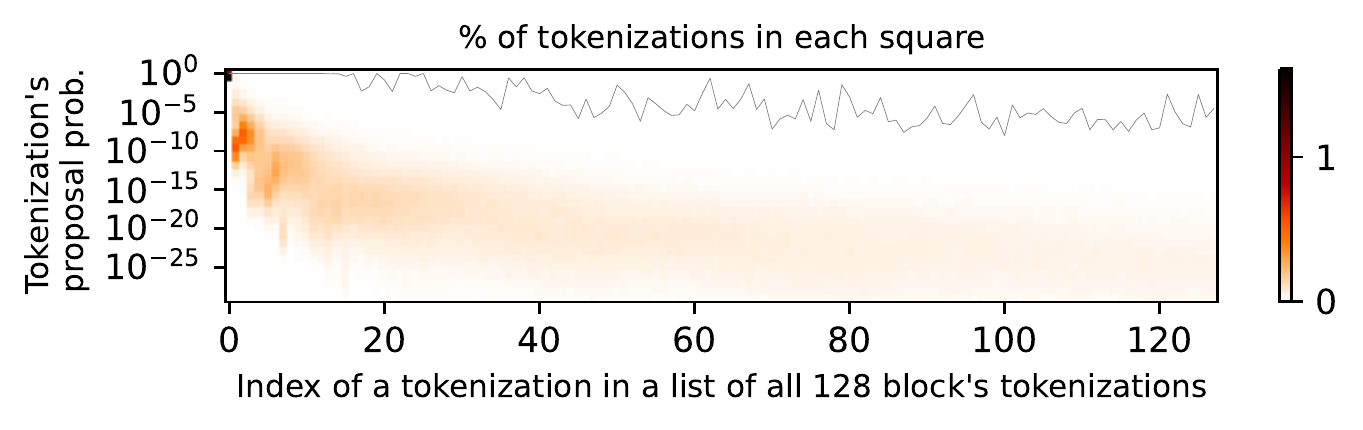}
        \caption{Top: the correlation between the number of subtokens in a block's tokenization and tokenization's proporal probability. Bottom: visualization of the proposal probabilities of blocks' tokenizations versus their ranks by the number of subtokens. Gray line specifies the maximum probability seen for each rank, which does not exceed $10^{-2}$ for ranks greater than 80. Both plots for GPT-2 on English Flores data.}
        \label{fig:maxtoksperblock}
\end{figure}

The proposed importance sampling algorithm limits $M$, the number of tokenizations per block which are scored with LM, for better efficiency. In this section we motivate why it is not harmful for the results. In the top plot of Figure~\ref{fig:maxtoksperblock} we show that the proposal probability of a block's tokenization strongly correlates with the number of subtokens in the tokenization. This motivates selecting top-$M$ tokenizations per block by sorting the block's tokenizations by the decreasing number of subtokens (we use $M=128$ in our experiments). Now, in the bottom plot of Figure~\ref{fig:maxtoksperblock} we present the 2d-histogram of proposal probabilities of blocks' tokenizations and their ranks in the sorting. It can be seen that the proposal probabilities of tokenizations with ranks higher than 10 have very low probabilities, i.e. usually lower than $10^{-10}$. In fact, tokenizations with ranks greater than 70 were never sampled (in 99.95\% one of the first 10 tokenizations was sampled, in 0.05\% cases --- one of tokenizations with indices 11--40, and in 0.0004\% cases --- with ranks 40--69.).

\section{Algorithm validation on short sentences}
\label{app:valid}
\begin{table}[t!]
\centering
\begin{tabular}{p{1.4cm}|p{1.4cm}|p{1.4cm}|p{1.4cm}} 
\toprule
$\BPCst-\BPCis$ & $\BPCis-\BPCm$ & $\BPCisL-\BPCm$ & $\BPCisR-\BPCm$ \\ \midrule
\multicolumn{4}{c}{N1: ``runspiration from quotes''} \\ \hline
.036271 & -.000479 & -.001765 & .000529 \\ \hline
\multicolumn{4}{c}{N2: ``Did organgatuangs fly''} \\ \hline
.023346 & .000675 & -.000129 & .001397 \\ \hline
\multicolumn{4}{c}{N3: ``the Buffalo-Pitt road''} \\ \hline
.000530 & .000313 & -.000293 & .000773 \\ \hline
\multicolumn{4}{c}{N4: ``It's Friday today''} \\ \hline
.000009 & .000003 & .000003 & .000003 \\ \hline
\multicolumn{4}{c}{N5: ``throughput of 600Mbit/s''} \\ \hline
.000170 & -.000001 & -.000001 & -.000001 \\ \hline
\multicolumn{4}{c}{N6: ``Television morning show''} \\ \hline
.000304 & -.000024 & -.000024 & -.000024 \\ \hline
\multicolumn{4}{c}{N7: ``snowboarding is cool''} \\ \hline
-.000136 & .000219 & .000038 & .0004613 \\ 
\bottomrule
\end{tabular}
\caption{Comparison of the proposed algorithm and exact marginalization for short sentences. We expect the value in the first column to be positive, the second value to be close to 0, the third value to be close to 0 and negative and the fourth value to be close to 0 and
positive (the last two conditions mean the conf. interval on $\BPCis$ includes $\BPCm$).}
\label{tab:enum}
\end{table}

To validate the proposed algorithm, we compare the marginal BPC estimated with our algorithm to the true marginal BPC, $\BPCm$, obtained by enumerating all tokenizations of several relatively short sentences ($\leqslant 25$ characters, $<$ 1M tokenizations). From Table~\ref{tab:enum} we observe that for sentences with relatively high BPC gap (N1--2), our estimate $\BPCis$ is close to $\BPCm$, with a thin confidence interval which includes $\BPCm$.
N3 shows the case with lower BPC gap, for which our estimate $\BPCis$ is 
between $\BPCst$ and $\BPCm$, and the confidence interval is wider but still includes $\BPCis$.
N4--6 show the case of low BPC gap, in which our proposal always sampled the 
default
tokenization hence 
there is no variance.
In all three cases the difference between our estimate ($\BPCis$) and the marginal ($\BPCm$) is 3--100 times smaller than between 
default
($\BPCst$) and marginal ($\BPCm$).
Finally, N7 shows the case with low BPC gap, in which our proposal did sample some non-default tokenizations, and the resulting estimate $\BPCis$ was larger than $\BPCst$. 
However, this ordering almost never happens with long texts, which is the intended use-case of our algorithm. To summarize, in almost all cases our algorithm produces a quite precise estimate.

\section{Additional confidence interval plots}
\label{app:confints}
\begin{figure}[t!]
    \centering
        \centering
        \includegraphics[width=\linewidth]{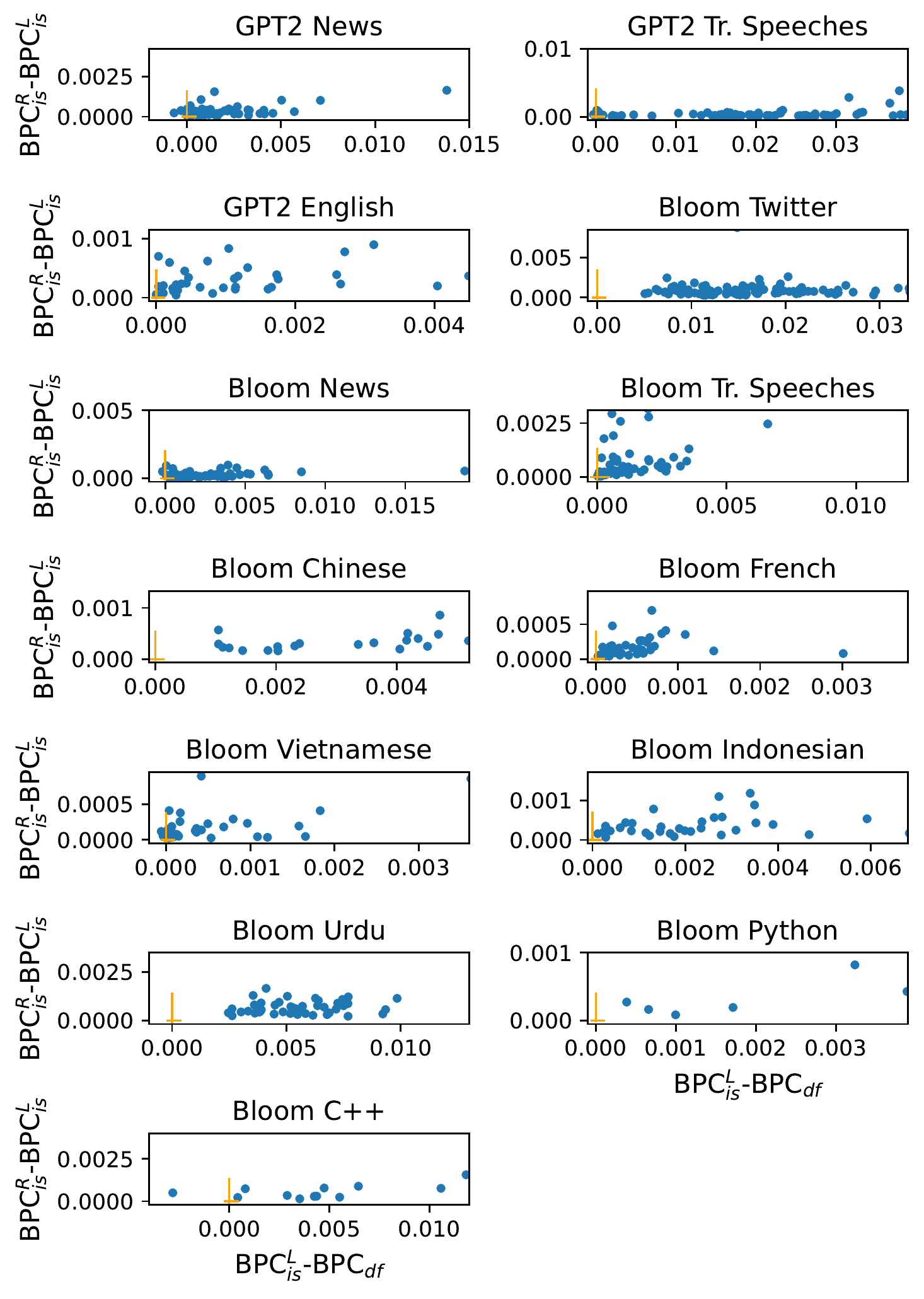}
        \caption{Confidence intervals visualisation. Each dot represents one data point (sequence). Please mind different axes scales.}
        \label{fig:confints_app}
\end{figure}

Figure~\ref{fig:confints_app} shows additional confidence interval plots. The conclusions are the same as for plots in the main text.

\section{Additional analysis}
\label{app:add_analysis}
The intuition why the impact of non-default tokenizations becomes more pronounced for complex words, low-resource languages and data distribution shift is that all these cases are characterized by the appearance of blocks which were rarely or never seen during training. Roughly speaking, frequent words are encoded with short token sequences (1-2 tokens) by design of the tokenizer. Furthermore, the language model assigns high probability to the default tokenizations of these words because it saw them frequently during training. As a result, the effect of marginalization is small. In contrast, rare words are encoded with longer token sequences, and because they are not frequently seen during training, the language model can assign high probabilities to other tokenizations than a default one. 

To illustrate given reasoning, Table~\ref{tab:additional} reports the distribution of number of tokens in sampled block’s tokenization, for low-frequency and high-frequency blocks, for GPT-2 on Twitter data. Low-frequency blocks are split into more tokens and have a higher portion of non-default tokenizations.

\begin{table}[t!]
\centering
\begin{small}
\begin{tabular}{c|cc} 
\toprule
\textbf{Block’s frequency} & \textbf{>=1e-4}  &  \textbf{<1e-4} \\ \midrule
\% such blocks                     & 0.602  & 0.398  \\ \midrule
\% sampled default tokenizations   & 0.978  & 0.925 \\
\% sampled non-default tokenizations   & 0.022  & 0.075 \\ \midrule
\% sampled length-1 tokenizations$^*$  & 0.829 & 0.306 \\
\% sampled length-2 tokenizations$^*$  & 0.137  & 0.299 \\
\% sampled length>=3 tokenizations$^*$ & 0.034 & 0.395  \\
\bottomrule
\end{tabular}
\end{small}
\caption{Additional analysis: the distribution of number of tokens in sampled block’s tokenization, for low-frequency and high-frequency blocks, for GPT-2 on Twitter data. Rows denoted with $^*$ include both default and non-default tokenizations.}
\label{tab:additional}
\end{table}

\end{document}

%% file: aux.tex
\newcommand{\Ptheta}{P}
\newcommand{\BPC}{\mathrm{BPC}}
\newcommand{\BPCst}{\mathrm{BPC}_{\mathrm{df}}}
\newcommand{\BPCis}{\mathrm{BPC}_{\mathrm{is}}}
\newcommand{\BPCisL}{\mathrm{BPC}^L_{\mathrm{is}}}
\newcommand{\BPCisR}{\mathrm{BPC}^R_{\mathrm{is}}}
\newcommand{\BPCm}{\mathrm{BPC}_{\mathrm{m}}}

\newif\ifready

\ifready
\newcommand{\gknote}[1]{}
\newcommand{\ncnote}[1]{}
\newcommand{\jrnote}[1]{}

\newcommand{\ptjr}[1]{#1}
\newcommand{\fmd}[1]{}

\else
\newcommand{\gknote}[1]{{\color{orange}\footnote{\textcolor{orange}{GK: #1}}}}
\newcommand{\ncnote}[1]{{\color{teal}\footnote{\textcolor{teal}{NC: #1}}}}
\newcommand{\jrnote}[1]{{\color{purple}\footnote{\textcolor{purple}{JR: #1}}}}

\newcommand{\ptjr}[1]{\textcolor{purple}{#1}}
\newcommand{\fmd}[1]{{\color{blue}\footnote{\textcolor{blue}{MD: #1}}}}

\fi